\documentclass[conference]{IEEEtran}
\IEEEoverridecommandlockouts
\usepackage{cite}
\usepackage{amsmath,amssymb,amsfonts}
\usepackage{graphicx}
\usepackage{textcomp}
\usepackage{xcolor}
\usepackage{multirow}
\usepackage{multicol}
\usepackage{booktabs}
\usepackage{hyperref}
\usepackage{algpseudocode}
\usepackage{color}
\usepackage{caption}
\usepackage{subcaption}
\usepackage[inline]{enumitem}
\usepackage{ulem}
\usepackage{makecell}

\def\BibTeX{{\rm B\kern-.05em{\sc i\kern-.025em b}\kern-.08em
    T\kern-.1667em\lower.7ex\hbox{E}\kern-.125emX}}
\begin{document}

\title{Transfer Risk Map: Mitigating Pixel-level Negative Transfer in Medical Segmentation \\
\thanks{\IEEEauthorrefmark{2} Corresponding author. 
}
}


\author{\IEEEauthorblockN{Shutong Duan, Jingyun Yang, Yang Tan, Guoqing Zhang, Yang Li\IEEEauthorrefmark{2}, Xiao-Ping Zhang}
 \IEEEauthorblockA{\textit{Shenzhen Key Laboratory of Ubiquitous Data Enabling}  \\
\textit{Tsinghua Shenzhen International Graduate School, Tsinghua University}  \\
  yangli@sz.tsinghua.edu.cn
  }
  }

\maketitle

\begin{abstract}
How to mitigate negative transfer in transfer learning is a long-standing and challenging issue, especially in the application of medical image segmentation. Existing methods for reducing negative transfer focuses on classification or regression tasks, ignoring the non-uniform  negative transfer risk in different image regions.  
In this work, we propose a simple yet effective weighted fine-tuning method that directs the model’s attention towards regions with significant transfer risk for medical semantic segmentation. Specifically, we compute a transferability-guided transfer risk map to quantify the transfer hardness for each pixel and the potential risks of negative transfer. During the fine-tuning phase, we introduce a map-weighted loss function, normalized with image foreground size to counter class imbalance. 
Extensive experiments on brain segmentation datasets show our method significantly improves the target task performance, with gains of 4.37\% on FeTS2021 and 1.81\% on iSeg-2019, avoiding negative transfer across modalities and tasks. Meanwhile, a 2.9\% gain under a few-shot scenario validates the robustness of our approach.

\end{abstract}

\section{Introduction}
\label{sec:intro}

For medical segmentation tasks, acquiring ground-truth labels is challenging because  detailed annotation of large 3D images is both time-consuming and requires expert input\cite{medical_deeplearning1, medical_deeplearning2}. To compensate the lack of labeled training samples, transfer learning has become a common approach for medical semantic segmentation \cite{medical_transfer1, medical_transfer2}, 
leveraging the knowledge from a high resource domain to improve the performance on low-resource tasks\cite{pan2009survey}. 
However, the effectiveness of transfer learning is not guaranteed: low task and domain correlation can impair the target domain performance. This phenomenon is known as  {\it negative transfer}\cite{negative_transfer_2}. 
Due to the inherent heterogeneity in medical images, including differences in imaging modalities, contrast variations, and patient-specific anatomy\cite{medical_transfer1, medical_da},   negative transfer  is common in medical image segmentation\cite{negative-transfer}. 

Most existing works address negative transfer by selecting more relevant source domain\cite{yicong, jin2024cross}, reweighting the source samples\cite{instance_level_3, negative_transfer_2}, and aligning distributions in the feature space\cite{feature_level_1}. The principle idea in these works lies in the notion of domain similarity \cite{MMD, KLD, correlation_coefficient} or some transferability metric across different tasks \cite{nce, hscore, leep, logme, otce} to quantify transfer performance.  
As these approaches have originated from a classification or regression context, they ignore the complex output structures in image-output tasks such as semantic segmentation. In segmentation, the risk of negative transfer of a given model does not uniformly distribute over the whole image. For example, some textures such as the white matter or the grey matter in the brain are shared across multiple domains, while some are unique to a specific domains such as the cortical protrusions and depressions across individual brains. This observation necessitates the need for a fine-grained transfer learning approach that could adapt to variable negative transfer risk at different image regions.

Recently, a few natural image segmentation works have incorporated the category-level and pixel-level transferability into loss function 
by propagating the global category-wise transferability calculated by entropy criterion\cite{dong}, or weighting the loss function by transferability score during fine-tuning phase\cite{10222912}. However, they don't work effectively on medical segmentation tasks due to fewer semantic categories, tiny segmentation foreground, and severed class imbalance of medical images, compared to natural images\cite{medical_and_natural}.

\begin{figure}
    \vspace{-0.5cm}
    \centering
    \includegraphics[width=\linewidth]{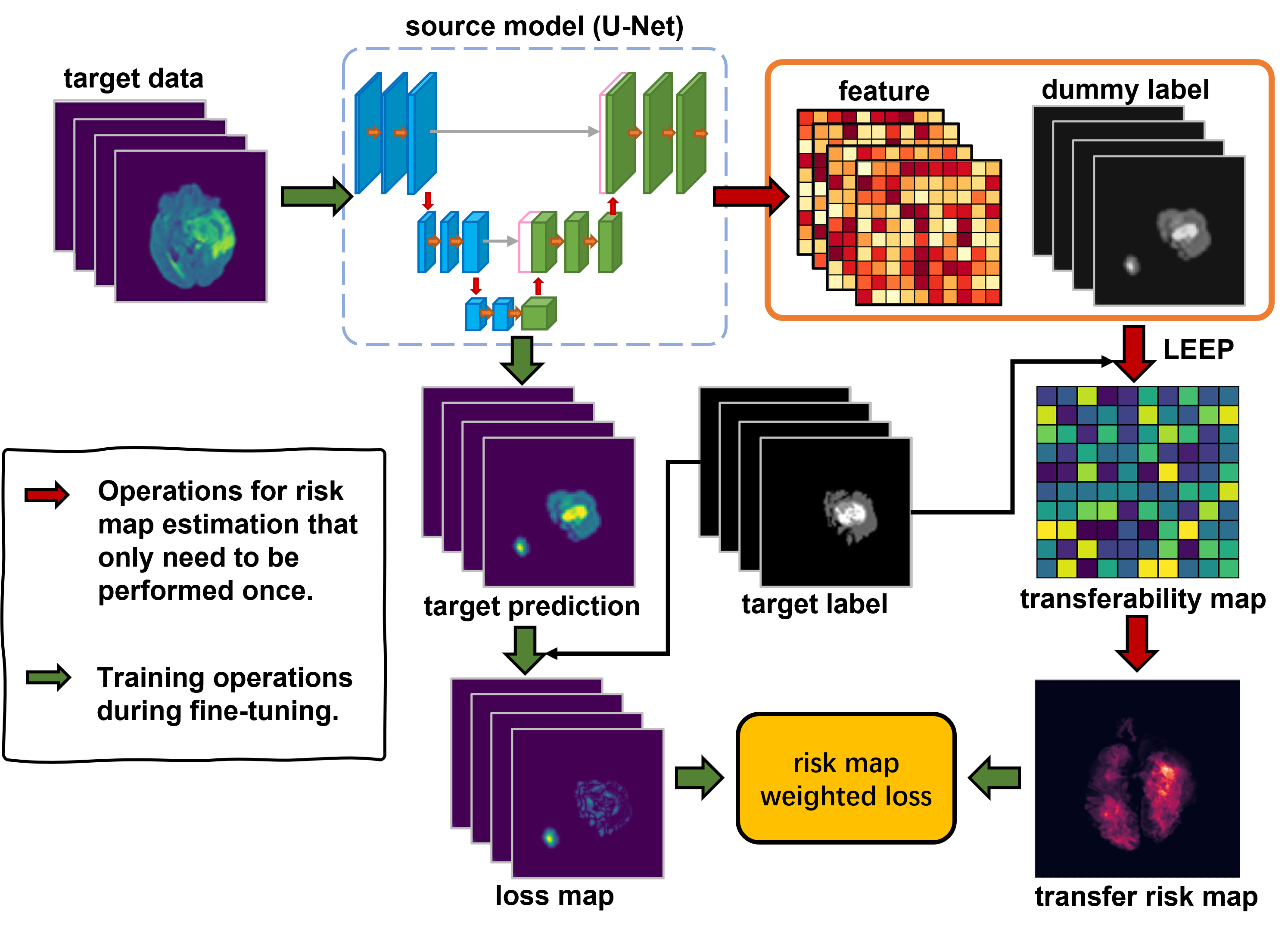}
    \caption{Illustration of the proposed transfer risk map guided weighted fine-tuning framework. 
    The strategy consists of: 1) a transferability-based pixel-level transfer risk map, which quantifies the potential risks of negative transfer for each pixel, 2) a transfer risk map weighted fine-tuning process, which allows the model's attention towards regions with significant transfer hardness.
    }
    \label{architecture}
    \vspace{-0.5cm}
\end{figure}

To solve the aforementioned challenges, we propose a simple yet effective weighted fine-tuning approach that directs the model's attention towards regions with significant transfer risk, tailored to the medical semantic segmentation problem (Fig.~\ref{architecture}). Specifically, we introduce a pixel-level {\it transfer risk map}, which quantifies the transfer hardness for each pixel and the potential risks of negative transfer associated with them. Here we adopt LEEP (Log Expected Empirical Prediction)\cite{leep} metric as our transferability method for its high computational efficiency, simplicity, and superior performance\cite{10222912}. And to alleviate the adverse effects of class imbalance during the fine-tuning phase, we calculate loss values across all pixels but average them exclusively over the foreground pixels. Such a method effectively reduces the impact that a large amount of well-learned background pixels might have on biasing the loss value, which in turn expedites the refinement of model parameters.

Extensive experiments on brain tumor and brain matter segmentation datasets demonstrate that our proposed fine-tuning method achieves significantly enhanced performance when transferring knowledge between distinct modalities and tasks, with a 4.37\% gain in brain tumor segmentation dataset FeTS 2021 and a 1.81\% gain in brain matter segmentation dataset iSeg-2019, indicating that it indeed avoids negative transfer from diverse modalities and tasks while learning beneficial knowledge for segmentation across multiple modalities and tasks. In the few-shot scenario, our method also improved the baseline by 2.9\% on average, validating the robustness of our approach under different sample sizes.  

\section{Methodology}
\subsection{Problem Definition}
In a typical source free transfer learning setting, We transfer the domain knowledge with only a pre-trained source model $\theta_S$ and without access to source data. Specifically, we are given a pre-trained model $\theta_S$ corresponding to a source task $S$ with $\mathcal{N}_S$ labeled samples $\{x_s^i,y_s^i \}_{i=1}^{\mathcal{N}_S}$ where $x_s^i \in \mathcal{X}_S, y_s^i \in \mathcal{Y}_S$, and also a target task $T$ with $\mathcal{N}_T$ labeled samples $\{x_t^i,y_t^i \}_{i=1}^{\mathcal{N}_T}$ where $x_t^i \in \mathcal{X}_T, y_t^i \in \mathcal{Y}_T$. $\mathcal{X}_S$, $\mathcal{X}_T$ represent the input image and $\mathcal{Y}_S$, $\mathcal{Y}_T$ represent their corresponding labels. The problem goal is to transfer the knowledge learned from the source to target that mitigate the negative transfer and improve the performance within the target domain.

Transferability reveals how easy it is to transfer knowledge learned from a source task to a target task. We can obtain the transferability from the source task to the target task by using the testing data of the target task $(\mathcal{X}_T, \mathcal{Y}_T)$ to calculate the expected log-likelihood of the source pretrained model $\theta_S$:
\begin{equation}
\label{Trans}
    Trs(\theta_S,\mathcal{X}_T,\mathcal{Y}_T) = \underset{x_i,y_i \in \mathcal{X}_T \times \mathcal{Y}_T}{\mathbb{E}} [logP( y_i\mid x_i;\theta_S)]
\end{equation}
\subsection{Transferability Estimation on Segmentation Task}
LEEP score is a more general metric for classification task, which is defined by the average log-likelihood of the expected empirical predictor, which predicts the dummy label distributions for target data in source label space and then computes the empirical conditional distribution of target label given the dummy source label. We select LEEP for our transferability estimation metric due to its high computational efficiency, simplicity and superior performance. We define the LEEP method as follows:
\begin{equation}
\label{LEEP}
    \operatorname{LEEP}(\theta_S,\mathcal{D}) = \frac{1}{n}\displaystyle\sum_{i = 1}^{n}\log(\displaystyle\sum_{z \in \mathcal{Z}}\hat{P}(y_i\mid z) \theta_S(x_i))
\end{equation}
where $\theta_S$ and $\mathcal{D} = \{(x_1, y_1),\cdots,(x_n, y_n)\}$ denote the source model and target dataset, respectively. $\mathcal{Z}$ denotes the dummy source label, $\hat{P}(\cdot\vert\cdot)$ denotes the empirical conditional distribution.

Then we adapt the LEEP score to work on medical segmentation task. The global feature vector extracted by source pretrained model can be precisely decomposed into distinct pixel-wise feature, each indexed according to its specific pixel coordinates. Subsequently, classification will be conducted on each pixel in input image, and output label will also be obtained at pixel-level, and transferability scores can be computed over the pixel-wise features treating each pixel as an instance of classifying. Now, pixel-level LEEP score for segmentation task can be defined as,
\begin{equation}
\label{LEEP-trans}
\resizebox{\columnwidth}{!}{%
    $\displaystyle 
    \operatorname{LEEP}(\theta_S,\mathcal{D}) = 
    \frac{1}{n} \sum_{i = 1}^{n} \left[ 
        \log \prod_{j = 1}^{W} \prod_{k = 1}^{H} \sum_{z \in \mathcal{Z}} 
        \hat{P}(y_i^{j,k} \mid z^{j,k}) \theta_S(x_i^{j,k})
    \right]
    $
}
\end{equation}
where $x_i \in \mathbb{R}^{W\times H\times1}$, $y_i \in \{0,1\}^{W\times H\times C_t}$. Here $W$, $H$ denote the width and height of image, and $C_t$ represent the number of categories of the target task.
\subsection{Transferability Guided Transfer Risk Weighted Fine-tuning}
After adapting the transferability metric to medical segmentation task, we can obtain pixel-level transferability, which characterizes the hardness of transferring knowledge at a local, granular level. Based on pixel-level transferability, we compute a transfer risk map that quantifies the potential risks of negative transfer associated with them. Furthermore, we propose a novel weighted fine-tuning approach guided by the transfer risk map. 

In particular, we define a pixel-level transferability map $t \in \mathbb{R}^{W\times H}$, and $t^{j,k}$ represents the transferability score at a pixel coordinate $(j,k)$, where $j \in [1,W]$, $k \in [1,H]$. Formally,
\begin{equation}
\label{transferability map}
   t^{j,k} = \operatorname{LEEP}(\theta_S,\{x_i^{j,k},y_i^{j,k} \}_{i=1}^{N}).
\end{equation}
Subsequently, we adjust the transferability map to scale within the interval of $[0, 1]$, with values approaching 1 representing pixels of higher transfer hardness and values near 0 denoting pixels with lesser transfer hardness. To further augment the model's emphasis on regions with increased transfer difficulty, we implement an exponential scaling with a base of 10 to introduce some nonlinearity. The transfer risk map $w \in \mathbb{R}^{W\times H}$ is defined as
\begin{equation}
\label{standerd transferability map}
   t_s^{j,k} = \frac{t^{j,k}-\min(t)}{\max(t)-\min(t)}
\end{equation}
\begin{equation}
\label{transferability risk map}
    w^{j,k} = 10^{t_s^{j,k}}.
\end{equation}
Then, we can define the transfer risk weighted loss, where
\begin{equation}
\label{transferability risk loss}
   L = \frac{\sum_{i = 1}^{N}\sum_{j = 1}^{W}\sum_{k = 1}^{H}w^{j,k}loss(\theta_S(x_i^{j,k},y_i^{j,k}))}{\sum_{i = 1}^{N}\sum_{j = 1}^{W}\sum_{k = 1}^{H}1\{y_i^{j,k} \not= 0\}}.
\end{equation}
It is important to highlight that we calculate loss values across all pixels but average them exclusively over the foreground pixels to alleviate the adverse effects of class imbalance during the fine-tuning phase. Such a method effectively prevents well-learned background pixels from diluting the loss value.

\section{Experiments and Results}
\subsection{Datasets}
We perform experiments on two publicly available brain MRI segmentation datasets: FeTS 2021 \cite{FeTS1,FeTS2,FeTS3} for brain tumor segmentation and iSeg-2019\cite{iseg} for brain matter segmentation. 
For each sample in FeTS 2021, volumes of 3 modalities are used, including T1-weighted (T1), T2-weighted (T2) and Fluid-Attenuated Inversion Recovery (FLAIR). The volume size is $240 \times 240 \times 155$. Corresponding labels of edematous tissue (ED), enhancing tumor (ET), and necrotic tumor core (NCR) are manually segmented by clinical experts. This dataset is further split into 22 partitions by the provider, according to different institutions and information extracted from images. Thus, each partition can be seen as an individual domain. For each sample in iSeg-2019 dataset, volumes of 2 modalities are available, including T1 and T2. The volume size is $144 \times 192 \times 256$. Corresponding labels of white matter (WM), gray matter (GM), and cerebrospinal fluid (CSF) are manually segmented by clinicians. 

To provide sufficient tasks for experimental analysis, we reorganize the iSeg-2019 dataset into a collection of binary segmentation tasks on every available modality. We perform cross-modality and cross-center four-category segmentation on the FeTS 2021 dataset, and cross-modality and cross-task binary segmentation on the iSeg-2019 dataset.

\subsection{Training Details}
Since the goal of this work is to demonstrate the effectiveness of transferability-guided transfer risk map on mitigating the potential local negative transfer during the fine-tuning process rather than striving for cutting-edge performance on medical image segmentation tasks, we employ the same model architecture for all experiments presented in this paper, a classic 2D U-Net\cite{unet}.  
For a fair comparison with all methods, we use the same input size, pre-processing strategy, and training loss for all methods. We adopt an Adam optimizer with a learning rate of 0.0001 to fine-tune the source model on the target data for 5k iterations. As for the transfer learning strategy, we follow the most common way which is pre-training the model on a source task and fine-tuning it on a target task. During the fine-tuning stage, the encoder is frozen and only the parameters of the decoder are updated. 
\begin{table}
	\centering
\centering
\setlength\tabcolsep{2pt}
\footnotesize
\caption{Dice score of vanilla fine-tuning, class weighted fine-tuning\cite{class-weight}, transferability map weighted fine-tuning\cite{10222912} and our transfer risk weighted fine-tuning on the iSeg-2019 dataset. Bold number: best score.}
\label{result1}
\resizebox{\columnwidth}{!}{%
\begin{tabular}{c|ccccccc|c}
\toprule
\multirow{2}{*}{{\bf \makecell[c]{Target\\ Task}} } & \multirow{2}{*}{{\bf Method} } & \multicolumn{6}{c}{{\bf Source Task}} & \multirow{2}{*}{{\bf \makecell[c]{Average\\ Dice}} }\\
\cmidrule(lr){3-8}
 & & CSF-T1 & GM-T1 & WM-T1 & CSF-T2 & GM-T2 & WM-T2 \\
\cmidrule{1-9}
\multirow{4}{*}{CSF-T1} & vanilla & \multirow{4}{*}{\textbackslash} & 0.9473 & 0.9406 & 0.9399 & 0.9354 & 0.9334 & 0.9393 \\ 
 & class weight &  & 0.9493 & 0.9487 & 0.9398 & 0.9352 & 0.9415 & 0.9429\\
 & Trs map &  & 0.9477 & 0.9410 & 0.9411 & 0.9355 & 0.9334 & 0.9397\\
 & ours &  & {\bf 0.9520} & {\bf 0.9502} & {\bf 0.9446} & {\bf 0.9432} & {\bf 0.9462} & {\bf 0.9472} \\
\cmidrule{1-9}
\multirow{4}{*}{GM-T1} & vanilla & 0.8972 & \multirow{4}{*}{\textbackslash} & 0.9010 & 0.8739 &  0.8858 & 0.8807 & 0.8877 \\
 & class weight & 0.8970 &  & 0.9006 & 0.8883 & 0.8979 & 0.8997 & 0.8967\\
 & Trs map & 0.8977 &  & 0.9010 & 0.8761 & 0.8848 & 0.8808 & 0.8881\\
 & ours & {\bf 0.9125} &  & {\bf 0.9107} & {\bf 0.9018} & {\bf 0.9001} & {\bf 0.9026} & {\bf 0.9055} \\
\cmidrule{1-9}
\multirow{4}{*}{WM-T1} & vanilla & 0.8787 & 0.8867 & \multirow{4}{*}{\textbackslash} & 0.8612 &  0.8560 & 0.8588 & 0.8683\\
 & class weight & 0.8831 & 0.8892 &  & 0.8707 & 0.8624 & 0.8642 & 0.8739\\
 & Trs map & 0.8787 & 0.8870 &  & 0.8604 & 0.8563 & 0.8596 & 0.8684\\
 & ours & {\bf 0.8896} & {\bf 0.8926} &  & {\bf 0.8755} & {\bf 0.8703} & {\bf 0.8782} & {\bf 0.8812} \\
\cmidrule{1-9}
\multirow{4}{*}{CSF-T2} & vanilla & 0.9229 & 0.9179 & 0.9143 & \multirow{4}{*}{\textbackslash} & 0.9210 & 0.9149 & 0.9182\\
 & class weight & 0.9260 & 0.9193 & 0.9133 &  & 0.9234 & 0.9145 & 0.9193\\
 & Trs map & 0.9233 & 0.9208 & 0.9163 &  & 0.9205 & 0.9150 & 0.9192\\
 & ours & {\bf 0.9319} & {\bf 0.9315} & {\bf 0.9321} &  & {\bf 0.9312} & {\bf 0.9329} & {\bf 0.9319}\\
\cmidrule{1-9}
\multirow{4}{*}{GM-T2} & vanilla & 0.8743 & 0.8641 & 0.8609 & 0.8674 & \multirow{4}{*}{\textbackslash} & 0.8810 & 0.8695\\
 & class weight & 0.8774 & 0.8686 & 0.8686 & 0.8716 &  & 0.8829 & 0.8738\\
 & Trs map & 0.8766 & 0.8695 & 0.8612 & 0.8674 &  & 0.8815 & 0.8712\\
 & ours & {\bf 0.8881} & {\bf 0.8846} & {\bf 0.8815} & {\bf 0.8851} &  & {\bf 0.8925} & {\bf 0.8864}\\
\cmidrule{1-9}
\multirow{4}{*}{WM-T2} & vanilla & 0.8486 & 0.8243 & 0.8243 & 0.8391 & 0.8389 & \multirow{4}{*}{\textbackslash} & 0.8350\\
 & class weight & 0.8361 & 0.8098 & 0.8298 & 0.8494 & 0.8511 &  & 0.8352\\
 & Trs map & 0.8497 & 0.8251 & 0.8279 & 0.8410 & 0.8471 &  & 0.8382\\
 & ours & {\bf 0.8630} & {\bf 0.8592} & {\bf 0.8582} & {\bf 0.8595} & {\bf 0.8676} &  & {\bf 0.8615}\\
\bottomrule
\end{tabular}
}
\vspace{-0.4cm}
\end{table}
\subsection{Evaluation on Medical Image Segmentation Task}
We apply our transferability-guided transfer risk weighted fine-tuning method on medical image segmentation tasks and compare the proposed method with the commonly used vanilla fine-tuning, class weighted fine-tuning\cite{class-weight}, and transferability map weighted fine-tuning\cite{10222912}. 

The iSeg-2019 dataset for brain matter segmentation closely resembles natural image segmentation tasks, exhibiting larger and more contiguous segmented regions, which demand the model's coarse-grained capability for object shape recognition. The FeTS 2021 dataset for brain tumor segmentation encompasses three distinct pathological labels that are characterized by small, discontinuous regions with hierarchical containment relationships. This demands a model endowed with a heightened fine-granularity for discerning local textures.

For iSeg-2019 dataset, quantitative comparisons shown in Table~\ref{result1} demonstrate that our proposed fine-tuning consistently outperforms the vanilla fine-tuning, class weighted fine-tuning and transferability map weighted fine-tuning in all transfer experiments. This approach can result in a notable increase in the Dice coefficient, reaching up to 3.49\%, with an average enhancement of 1.6\% and an overall gain of 1.81\%. 

For FeTS 2021 dataset, as shown in Table~\ref{result2}, our proposed fine-tuning method surpasses vanilla fine-tuning, class weighted fine-tuning and transferability map weighted fine-tuning across a spectrum of transfer experiments. It achieves a remarkable maximum increase of 8.3\% in the Dice coefficient, with an average enhancement of 2\% and an overall gain of 4.37\%.
\begin{table}
	\centering
\centering
\setlength\tabcolsep{2pt}
\footnotesize
\caption{Dice score of vanilla fine-tuning, class weighted fine-tuning\cite{class-weight}, transferability map weighted fine-tuning\cite{10222912} and our transfer risk weighted fine-tuning on the FeTS2021 dataset. Bold number: best score.}
\label{result2}
\resizebox{\columnwidth}{!}{%
\begin{tabular}{c|ccccccc|c}
\toprule
\multirow{2}{*}{{\bf \makecell[c]{Target\\ Task}} } & \multirow{2}{*}{{\bf Method} } & \multicolumn{6}{c}{{\bf Source Task}} & \multirow{2}{*}{{\bf \makecell[c]{Average \\Dice}} }\\
\cmidrule(lr){3-8}
 & & 13-T1 & 14-T1 & 14-T2 & 14-FLAIR & 17-T2 & 18-FLAIR \\
\cmidrule{1-9}
\multirow{4}{*}{13-T1} & vanilla  & \multirow{4}{*}{\textbackslash} & 0.422 & 0.304 & 0.352 & 0.317 & 0.303 & 0.340\\ 
 & class weight &  & {\bf 0.427} & 0.303 & 0.357 & 0.315 & 0.303 & 0.341\\
 & Trs map &  & 0.420 & 0.307 & 0.355 & 0.316 & 0.309 & 0.341\\
 & ours &  & 0.426 & {\bf 0.315} & {\bf 0.364} & {\bf 0.319} & {\bf 0.318} & {\bf 0.348}\\
\cmidrule{1-9}
\multirow{4}{*}{14-T1} & vanilla & 0.491 & \multirow{4}{*}{\textbackslash} & 0.457 & 0.364 &  0.410 & 0.451 & 0.435\\
 & class weight & 0.503 &  & 0.471 & 0.383 & 0.396 & 0.456 & 0.442\\
 & Trs map & 0.498 &  & 0.459 & 0.366 & 0.408 & 0.454 & 0.437\\
 & ours & {\bf 0.524} &  & {\bf 0.488} & {\bf 0.398} & {\bf 0.443} & {\bf 0.473} & {\bf 0.465}\\
\cmidrule{1-9}
\multirow{4}{*}{14-T2} & vanilla & 0.413 & 0.464 & \multirow{4}{*}{\textbackslash} & 0.464 &  0.508 & 0.520 & 0.474\\
 & class weight & 0.411 & 0.477 &  & 0.468 & 0.508 & {\bf 0.525} & 0.478\\
 & Trs map & 0.417 & 0.473 &  & 0.471 & 0.509 & 0.515 & 0.477\\
 & ours & {\bf 0.431} & {\bf 0.505} &  & {\bf 0.482} & {\bf 0.519} & 0.521 & {\bf 0.492}\\
\cmidrule{1-9}
\multirow{4}{*}{14-FLAIR} & vanilla & 0.484 & 0.521 & 0.582 & \multirow{4}{*}{\textbackslash} & 0.535 & 0.578 & 0.540\\
 & class weight & 0.479 & 0.529 & 0.588 &  & 0.532 & 0.586 & 0.543\\
 & Trs map & 0.487 & 0.522 & 0.586 &  & 0.533 & 0.578 & 0.541\\
 & ours & {\bf 0.504} & {\bf 0.559} & {\bf 0.590} &  & {\bf 0.536} & {\bf 0.588} & {\bf 0.555}\\
\cmidrule{1-9}
\multirow{4}{*}{17-T2} & vanilla & 0.313 & 0.365 & 0.393 & 0.313 & \multirow{4}{*}{\textbackslash} & 0.353 & 0.347\\
 & class weight & 0.316 & 0.366 & 0.389 & 0.327 &  & 0.352 & 0.350\\
 & Trs map & 0.315 & 0.368 & 0.392 & 0.315 &  & 0.356 & 0.349\\
 & ours & {\bf 0.320} & {\bf 0.384} & {\bf 0.408} & {\bf 0.396} &  & {\bf 0.365} & {\bf 0.375}\\
\cmidrule{1-9}
\multirow{4}{*}{18-FLAIR} & vanilla & 0.421 & 0.432 & 0.473 & 0.496 & 0.483 & \multirow{4}{*}{\textbackslash} & 0.461\\
 & class weight & 0.417 & 0.433 & 0.473 & 0.509 & 0.482 &  & 0.463\\
 & Trs map & 0.419 & 0.438 & 0.477 & 0.506 & 0.475 &  & 0.463\\
 & ours & {\bf 0.426} & {\bf 0.451} & {\bf 0.483} & {\bf 0.530} & {\bf 0.483} &  & {\bf 0.475}\\
\bottomrule
\end{tabular}
}
\end{table}
\begin{table}
	\centering
\centering
\setlength\tabcolsep{2pt}
\footnotesize
\caption{Dice score of transferability based transfer risk weighted fine-tuning and vanilla fine-tuning on the iSeg-2019 dataset under few-shot scenario. Bold number: best score.}
\label{result3}
\resizebox{\columnwidth}{!}{%
\begin{tabular}{c|cccccc|c}
\toprule
\multirow{2}{*}{{\bf \makecell[c]{Sample \\ Nums}}} & \multirow{2}{*}{{\bf Method} } & \multicolumn{5}{c}{{\bf Target Task}} & \multirow{2}{*}{{\bf \makecell[c]{Average\\ Dice}} }\\
\cmidrule(lr){3-7}
& & GM-T1 & WM-T1 & CSF-T2 & GM-T2 & WM-T2 & \\
\cmidrule{1-8}
\multirow{2}{*}{400} & vanilla & 0.8915 & 0.8608 & 0.9175 & 0.8592 & 0.8228 & 0.8704 \\ 
 & ours & {\bf 0.9072} & {\bf 0.8750} & {\bf 0.9299} & {\bf 0.8858} & {\bf 0.8498} & {\bf 0.8895} \\
\cmidrule{1-8}
\multirow{2}{*}{100} & vanilla & 0.8706 & 0.8480 & 0.9041 & 0.8443 & 0.8027 & 0.8539 \\
 & ours & {\bf 0.8925} & {\bf 0.8599} & {\bf 0.9223} & {\bf 0.8582} & {\bf 0.8273} & {\bf 0.8720} \\
\cmidrule{1-8}
\multirow{2}{*}{50} & vanilla & 0.8572 & 0.8488 & 0.8951 & 0.8170 & 0.7913 & 0.8419 \\
 & ours & {\bf 0.8927} & {\bf 0.8533} & {\bf 0.9113} & {\bf 0.8584} & {\bf 0.8139} & {\bf 0.8659} \\
\bottomrule
\end{tabular}
}
\vspace{-0.4cm}
\end{table}
\subsection{Study on Few- shot Scenarios}
We conducted experiments under few-shot scenarios on the iSeg-2019 dataset, testing the performance of our proposed method when the target task had only 400, 100, and 50 annotated 2d slice images, respectively. As demonstrated in the Table~\ref{result3}, employing one task as the source model, we conducted transfer fine-tuning on the remaining five tasks. Irrespective of the quantity of target images, our method consistently outperformed vanilla fine-tuning. Remarkably, even with a minimal training slice sample size of 50, our method still achieved a 2.4\% improvement and a 2.9\% gain in Dice score in average. These experimental findings validate the robustness of our approach in a few-shot context. 
\subsection{Effectiveness of Negative Transfer Mitigation}
\begin{figure}
    \vspace{-0.4cm}
    \centering
    \includegraphics[width=\linewidth]{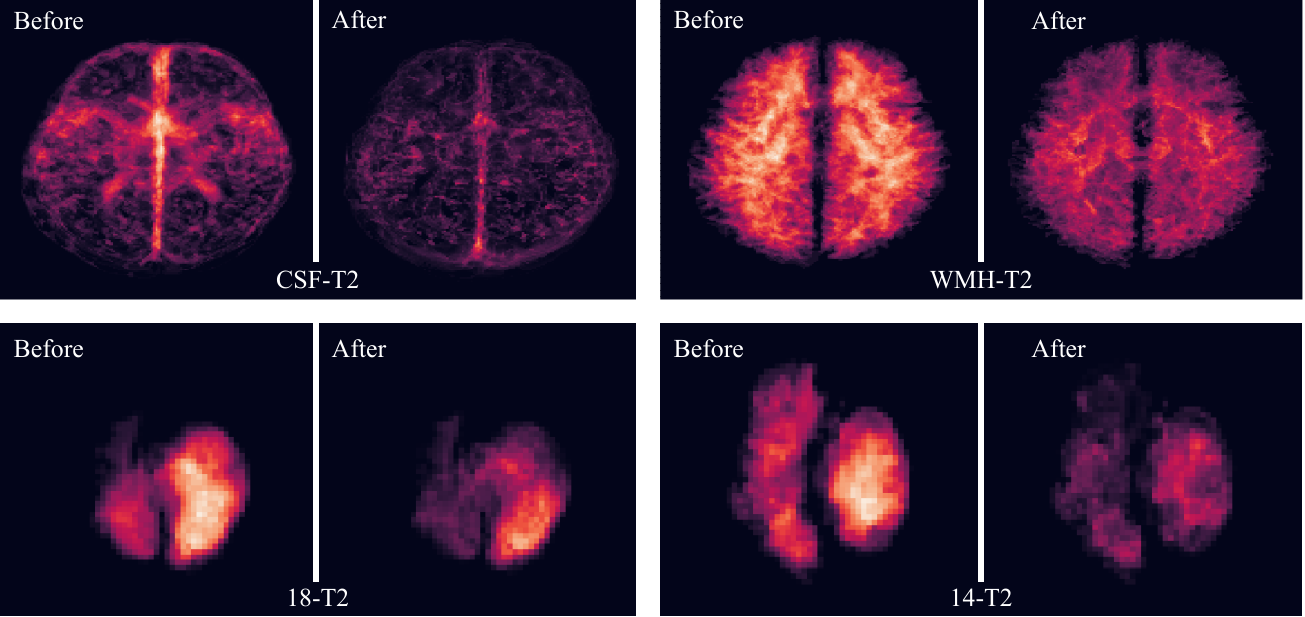}
    \caption{The pixel-level transfer risk maps before and after applying our fine-tuning method between the source model and the target data. The first row shows the results on iSeg-2019 with CSF-T1 as the source task, while the second row displays results on FeTS 2021 with 14-T1 as the source task.
    }
    \label{transferability}
    \vspace{-0.4cm}
\end{figure}

To investigate the actual impact of local transfer risk on the fine-tuning phase, we visualized the transfer risk maps before and after applying our fine-tuning method between the source model and the target data. As shown in Fig.~\ref{transferability}, it shows the transfer risk maps before and after the fine-tuning phase between the pre-trained source model and the target tasks in the iSeg-2019 and FeTS 2021 dataset. Darker colors indicate lower transfer risk, while brighter colors signify higher transfer risk. It is evident that our fine-tuning approach has effectively condensed the extensive regions of high transfer risk into well-defined, smaller segments, concurrently reducing the transfer hardness.
Additionally, we observed that regions exhibiting the most significant transfer hardness have undergone a transformation, from resembling segmentation contours of source task to the precise segmentation patterns of target tasks. The comparison of transfer risk maps pre- and post-fine-tuning validates the efficacy of our proposed approach in mitigating the risk of potential negative transfer. The simple weighting procedure not only reduces the overall transfer hardness but also refines the scope of potential negative transfer risk regions during model training, effectively redirecting the model's attention towards pertinent features.
\section{Conclusion}
In this work, we propose a simple yet effective weighted fine-tuning approach that directs the model's attention towards regions with significant transfer risk, tailored to the medical semantic segmentation problem. Specifically, we introduce a pixel-level transferability-guided transfer risk map, which quantifies the transfer hardness for each pixel and the potential risks of negative transfer associated with them. During the fine-tuning phase, we calculate weighted loss values across all pixels and average them exclusively over the foreground pixels. Extensive experiments demonstrate that our proposed fine-tuning method achieves significantly enhanced performance when transferring knowledge between distinct modalities and tasks, indicating that it indeed avoids negative transfer from diverse modalities and tasks while learning beneficial knowledge for segmentation across multiple modalities and tasks. The simplicity of the proposed method also makes it easy to integrate into more advanced architectures such as vision transformer, and more sophisticated transfer learning paradigms.

\section*{Acknowledgment}
This work was supported in part by the Natural Science Foundation of China (Grant 62371270), Shenzhen Key Laboratory of Ubiquitous Data Enabling (No.ZDSYS20220527171406015), and Tsinghua Shenzhen International Graduate School-Shenzhen Pengrui Endowed Professorship Scheme of Shenzhen Pengrui Foundation.

\bibliographystyle{IEEEtran}
\normalem
\bibliography{main}

\begin{thebibliography}{10}
\providecommand{\url}[1]{#1}
\csname url@samestyle\endcsname
\providecommand{\newblock}{\relax}
\providecommand{\bibinfo}[2]{#2}
\providecommand{\BIBentrySTDinterwordspacing}{\spaceskip=0pt\relax}
\providecommand{\BIBentryALTinterwordstretchfactor}{4}
\providecommand{\BIBentryALTinterwordspacing}{\spaceskip=\fontdimen2\font plus
\BIBentryALTinterwordstretchfactor\fontdimen3\font minus \fontdimen4\font\relax}
\providecommand{\BIBforeignlanguage}[2]{{%
\expandafter\ifx\csname l@#1\endcsname\relax
\typeout{** WARNING: IEEEtran.bst: No hyphenation pattern has been}%
\typeout{** loaded for the language `#1'. Using the pattern for}%
\typeout{** the default language instead.}%
\else
\language=\csname l@#1\endcsname
\fi
#2}}
\providecommand{\BIBdecl}{\relax}
\BIBdecl

\bibitem{medical_deeplearning1}
R.~Wang, T.~Lei, R.~Cui, B.~Zhang, H.~Meng, and A.~K. Nandi, ``Medical image segmentation using deep learning: A survey,'' \emph{IET image processing}, vol.~16, no.~5, pp. 1243--1267, 2022.

\bibitem{medical_deeplearning2}
M.~H. Hesamian, W.~Jia, X.~He, and P.~Kennedy, ``Deep learning techniques for medical image segmentation: achievements and challenges,'' \emph{Journal of digital imaging}, vol.~32, pp. 582--596, 2019.

\bibitem{medical_transfer1}
P.~Kora, C.~P. Ooi, O.~Faust, U.~Raghavendra, A.~Gudigar, W.~Y. Chan, K.~Meenakshi, K.~Swaraja, P.~Plawiak, and U.~R. Acharya, ``Transfer learning techniques for medical image analysis: A review,'' \emph{Biocybernetics and Biomedical Engineering}, vol.~42, no.~1, pp. 79--107, 2022.

\bibitem{medical_transfer2}
D.~Karimi, S.~K. Warfield, and A.~Gholipour, ``Transfer learning in medical image segmentation: New insights from analysis of the dynamics of model parameters and learned representations,'' \emph{Artificial intelligence in medicine}, vol. 116, p. 102078, 2021.

\bibitem{pan2009survey}
S.~J. Pan and Q.~Yang, ``A survey on transfer learning,'' \emph{IEEE Transactions on knowledge and data engineering}, vol.~22, no.~10, pp. 1345--1359, 2009.

\bibitem{negative_transfer_2}
Z.~Wang, Z.~Dai, B.~Poczos, and J.~Carbonell, ``Characterizing and avoiding negative transfer,'' in \emph{Proceedings of the IEEE/CVF Conference on Computer Vision and Pattern Recognition (CVPR)}, June 2019.

\bibitem{medical_da}
H.~Guan and M.~Liu, ``Domain adaptation for medical image analysis: a survey,'' \emph{IEEE Transactions on Biomedical Engineering}, vol.~69, no.~3, pp. 1173--1185, 2021.

\bibitem{negative-transfer}
W.~Zhang, L.~Deng, L.~Zhang, and D.~Wu, ``A survey on negative transfer,'' \emph{IEEE/CAA Journal of Automatica Sinica}, vol.~10, no.~2, pp. 305--329, 2023.

\bibitem{yicong}
Y.~Li, Y.~Tan, J.~Yang, Y.~Li, and X.-P. Zhang, ``Finding the most transferable tasks for brain image segmentation,'' in \emph{2022 IEEE International Conference on Bioinformatics and Biomedicine (BIBM)}.\hskip 1em plus 0.5em minus 0.4em\relax IEEE, 2022, pp. 1620--1625.

\bibitem{jin2024cross}
J.~Jin, G.~Bai, R.~Xu, K.~Qin, H.~Sun, X.~Wang, and A.~Cichocki, ``A cross-dataset adaptive domain selection transfer learning framework for motor imagery-based brain-computer interfaces,'' \emph{Journal of Neural Engineering}, vol.~21, no.~3, p. 036057, 2024.

\bibitem{instance_level_3}
Z.~Peng, Y.~Jia, and J.~Hou, ``Non-negative transfer learning with consistent inter-domain distribution,'' \emph{IEEE Signal Processing Letters}, vol.~27, pp. 1720--1724, 2020.

\bibitem{feature_level_1}
M.~Long, J.~Wang, G.~Ding, W.~Cheng, X.~Zhang, and W.~Wang, ``Dual transfer learning,'' in \emph{Proceedings of the 2012 SIAM International Conference on Data Mining}.\hskip 1em plus 0.5em minus 0.4em\relax SIAM, 2012, pp. 540--551.

\bibitem{MMD}
A.~Gretton, K.~M. Borgwardt, M.~J. Rasch, B.~Sch{\"o}lkopf, and A.~Smola, ``A kernel two-sample test,'' \emph{The Journal of Machine Learning Research}, vol.~13, no.~1, pp. 723--773, 2012.

\bibitem{KLD}
S.~Kullback and R.~A. Leibler, ``On information and sufficiency,'' \emph{The annals of mathematical statistics}, vol.~22, no.~1, pp. 79--86, 1951.

\bibitem{correlation_coefficient}
Y.-P. Lin and T.-P. Jung, ``Improving eeg-based emotion classification using conditional transfer learning,'' \emph{Frontiers in human neuroscience}, vol.~11, p. 334, 2017.

\bibitem{nce}
A.~T. Tran, C.~V. Nguyen, and T.~Hassner, ``Transferability and hardness of supervised classification tasks,'' in \emph{Proceedings of the IEEE/CVF International Conference on Computer Vision}, 2019, pp. 1395--1405.

\bibitem{hscore}
Y.~Bao, Y.~Li, S.-L. Huang, L.~Zhang, L.~Zheng, A.~Zamir, and L.~Guibas, ``An information-theoretic approach to transferability in task transfer learning,'' in \emph{2019 IEEE international conference on image processing (ICIP)}.\hskip 1em plus 0.5em minus 0.4em\relax IEEE, 2019, pp. 2309--2313.

\bibitem{leep}
C.~Nguyen, T.~Hassner, M.~Seeger, and C.~Archambeau, ``Leep: A new measure to evaluate transferability of learned representations,'' in \emph{International Conference on Machine Learning}.\hskip 1em plus 0.5em minus 0.4em\relax PMLR, 2020, pp. 7294--7305.

\bibitem{logme}
K.~You, Y.~Liu, J.~Wang, and M.~Long, ``Logme: Practical assessment of pre-trained models for transfer learning,'' in \emph{International Conference on Machine Learning}.\hskip 1em plus 0.5em minus 0.4em\relax PMLR, 2021, pp. 12\,133--12\,143.

\bibitem{otce}
Y.~Tan, Y.~Li, and S.-L. Huang, ``Otce: A transferability metric for cross-domain cross-task representations,'' in \emph{Proceedings of the IEEE/CVF conference on computer vision and pattern recognition}, 2021, pp. 15\,779--15\,788.

\bibitem{dong}
J.~Dong, Y.~Cong, G.~Sun, Z.~Fang, and Z.~Ding, ``Where and how to transfer: Knowledge aggregation-induced transferability perception for unsupervised domain adaptation,'' \emph{IEEE Transactions on Pattern Analysis and Machine Intelligence}, vol.~46, no.~3, pp. 1664--1681, 2024.

\bibitem{10222912}
Y.~Tan, Y.~Li, Y.~Li, and X.-P. Zhang, ``Efficient prediction of model transferability in semantic segmentation tasks,'' in \emph{2023 IEEE International Conference on Image Processing (ICIP)}, 2023, pp. 720--724.

\bibitem{medical_and_natural}
S.~Asgari~Taghanaki, K.~Abhishek, J.~P. Cohen, J.~Cohen-Adad, and G.~Hamarneh, ``Deep semantic segmentation of natural and medical images: a review,'' \emph{Artificial Intelligence Review}, vol.~54, pp. 137--178, 2021.

\bibitem{FeTS1}
S.~Pati, U.~Baid, M.~Zenk, B.~Edwards, M.~Sheller, G.~A. Reina, P.~Foley, A.~Gruzdev, J.~Martin, S.~Albarqouni \emph{et~al.}, ``The federated tumor segmentation (fets) challenge,'' \emph{arXiv preprint arXiv:2105.05874}, 2021.

\bibitem{FeTS2}
G.~A. Reina, A.~Gruzdev, P.~Foley, O.~Perepelkina, M.~Sharma, I.~Davidyuk, I.~Trushkin, M.~Radionov, A.~Mokrov, D.~Agapov \emph{et~al.}, ``Openfl: An open-source framework for federated learning,'' \emph{arXiv preprint arXiv:2105.06413}, 2021.

\bibitem{FeTS3}
S.~Bakas, H.~Akbari, A.~Sotiras, M.~Bilello, M.~Rozycki, J.~S. Kirby, J.~B. Freymann, K.~Farahani, and C.~Davatzikos, ``Advancing the cancer genome atlas glioma mri collections with expert segmentation labels and radiomic features,'' \emph{Scientific data}, vol.~4, no.~1, pp. 1--13, 2017.

\bibitem{iseg}
Y.~Sun, K.~Gao, Z.~Wu, G.~Li, X.~Zong, Z.~Lei, Y.~Wei, J.~Ma, X.~Yang, X.~Feng \emph{et~al.}, ``Multi-site infant brain segmentation algorithms: the iseg-2019 challenge,'' \emph{IEEE Transactions on Medical Imaging}, vol.~40, no.~5, pp. 1363--1376, 2021.

\bibitem{unet}
O.~Ronneberger, P.~Fischer, and T.~Brox, ``U-net: Convolutional networks for biomedical image segmentation,'' in \emph{Medical Image Computing and Computer-Assisted Intervention -- MICCAI 2015}, N.~Navab, J.~Hornegger, W.~M. Wells, and A.~F. Frangi, Eds.\hskip 1em plus 0.5em minus 0.4em\relax Cham: Springer International Publishing, 2015, pp. 234--241.

\bibitem{class-weight}
D.~N. Le, H.~X. Le, L.~T. Ngo, and H.~T. Ngo, ``Transfer learning with class-weighted and focal loss function for automatic skin cancer classification,'' \emph{arXiv preprint arXiv:2009.05977}, 2020.

\end{thebibliography}

\vspace{12pt}

\end{document}